\documentclass[conference]{IEEEtran}
\IEEEoverridecommandlockouts
\usepackage{cite}
\usepackage{amsmath,amssymb,amsfonts}
\usepackage{algorithmic}
\usepackage{graphicx}
\usepackage{textcomp}
\usepackage{xcolor}
\usepackage{tikz}
\usepackage{caption}
\usepackage{subcaption}
\usepackage{multicol}
\usepackage{siunitx}
\usetikzlibrary{patterns}
\usetikzlibrary{matrix}
\usepackage{pgfplots}
\usepackage{afterpage}
\pgfplotsset{compat=1.17}

\DeclareMathOperator*{\argmax}{arg\,max}
\DeclareMathOperator*{\argmin}{arg\,min}
\newcommand{\norm}[1]{\left\lVert#1\right\rVert}

\usepackage{todonotes}

\def\BibTeX{{\rm B\kern-.05em{\sc i\kern-.025em b}\kern-.08em
    T\kern-.1667em\lower.7ex\hbox{E}\kern-.125emX}}

\begin{document}

\title{FIT-SLAM - Fisher Information and Traversability estimation-based Active SLAM for exploration in 3D environments\\
\thanks{ This research was funded by Defense Innovation Agency (AID) of the French Ministry of Defense, Research Project CONCORDE No 2019 65 0090004707501.
This paper is a draft and unauthorized use of the presented work is strictly prohibited.}
}

\author{\IEEEauthorblockN{Suchetan Saravanan}
\IEEEauthorblockA{\textit{ISAE-SUPAERO}, \\
\textit{University of Toulouse}\\
Toulouse, France \\
\textit{BITS Pilani, India} \\
suchetan.saravanan@isae.fr}
\and
\IEEEauthorblockN{Corentin Chauffaut}
\IEEEauthorblockA{\textit{ISAE-SUPAERO}, \\
\textit{University of Toulouse}\\
Toulouse, France \\
corentin.chaffaut@isae.fr \\
}
\and
\IEEEauthorblockN{Caroline Chanel}
\IEEEauthorblockA{\textit{ISAE-SUPAERO}, \\
\textit{University of Toulouse}\\
Toulouse, France \\
caroline.chanel@isae.fr \\}
~\\
\and
\IEEEauthorblockN{Damien Vivet}
\IEEEauthorblockA{\textit{ISAE-SUPAERO}, \\
\textit{University of Toulouse}\\
Toulouse, France \\
damien.vivet@isae.fr \\
*Corresponding author
}
}

\maketitle

\begin{abstract} Active visual SLAM finds a wide array of applications in GNSS-Denied sub-terrain environments and outdoor environments for ground robots. To achieve robust localization and mapping accuracy, it is imperative to incorporate the perception considerations in the goal selection and path planning towards the goal during an exploration mission. Through this work, we propose FIT-SLAM (Fisher Information and Traversability estimation-based Active SLAM), a new exploration method tailored for unmanned ground vehicles (UGVs) to explore 3D environments. This approach is devised with the dual objectives of sustaining an efficient exploration rate while optimizing SLAM accuracy. Initially, an estimation of a global traversability map is conducted, which accounts for the environmental constraints pertaining to traversability. Subsequently, we propose a goal candidate selection approach along with a path planning method towards this goal that takes into account the information provided by the landmarks used by the SLAM backend to achieve robust localization and successful path execution . The entire algorithm is tested and evaluated first in a simulated 3D world, followed by a real-world environment and is compared to pre-existing exploration methods. The results obtained during this evaluation demonstrate a significant increase in the exploration rate while effectively minimizing the localization covariance.
\end{abstract}



\section{Introduction}
For mobile systems to be robustly localized, actively considering the perception requirement in the planning stage is
essential. On-board visual sensing and computing permit systems to operate autonomously but bring additional constraints
to motion planning algorithms. Specifically, the robot’s motion
impacts the information the visual sensors will capture and thus influences the performance of perception
and localization algorithms. Therefore, the requirement of
visual perception has to be taken into consideration in motion planning in order to improve localization accuracy. This
problem is known as active vision [1] or active perception [2].
The central paradigm of active simultaneous localization and
mapping (ASLAM) is to plan the sensor motion based on the
information that can be attained and the covariance that can
be maintained during the exploration mission.

The critical goal in UGV-based exploration is to achieve good
long-term mission planning that has a competitive exploration
rate while also maintaining good mapping and localization accuracy during the mission. The algorithms adeptly tackle the
exploration-exploitation dilemma. That is, to strike a balance
between exploring new parts of the environment and exploiting
the already explored portion of the map.


\begin{figure}[t]
    \input{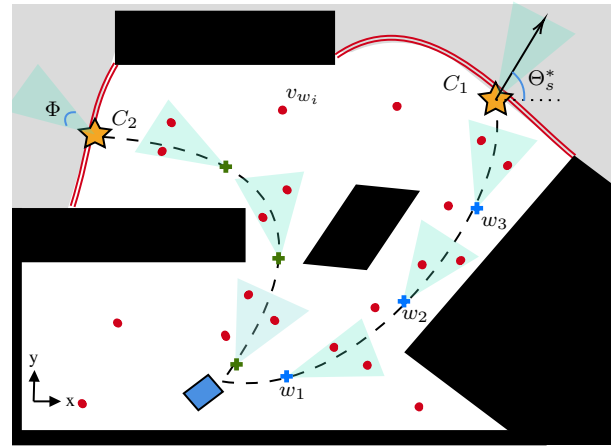}
    \caption{Overview of the proposed framework: A 2D thresholded traversability map is analyzed to extract frontiers goals (stars). Each goal candidate $C_i$ is ranked based on the information computed for the planned path. Such information is linked to the observed landmarks (red dots) in the camera field-of-view (FOV) $\Phi$ (blue area) but also the maximal reduction in entropy after reaching the goal with robot orientation $\Theta_s^*$. This allows to ensure good localizability while exploring new areas, resulting in a more accurate mapping.}
    \label{fig:enter-label}
\end{figure}

B. Yamuchi \cite{BYamuchi} introduced the concept of \textit{Frontiers}, a set of points which acts as the transition region between the areas of a map that are already explored from those not yet visited. Since this work, where a greedy \textit{frontier-based} exploration was performed, several contributions and improvements have been made to the exploration strategy. In the realm of goal determination for exploration, there exists a variety of methodologies. The frontier-based approach is still popular and can be found in recent works such as \cite{BONETTO2022104102}. An alternative method to select possible goals is proposed by Umari et al. \cite{rrtumari}, where they detect the frontier on a 2D occupancy grid map using rapidly-exploring random trees (RRTs). The use of this strategy concatenated with computer vision algorithms can be found in works such as \cite{rrtusage}. These approaches, however are limited to the robot exploring a 2D planar environment, which vastly reduces the capabilities of a UGV.
Some works in the past focus on 3D UGV exploration, such as \cite{RobotIntractedTraversability, AhmadSamplingSubT, 3DExplorationSelin}. These works showcase promising results in terms of the exploration rate of the proposed algorithm. However, they do not account for the localization accuracy estimated by the SLAM backend.

Unfortunately, high uncertainty in the robot state could lead to a significantly unreliable map. In \cite{seminalbourgalt}, Bourgault et al. addressed this issue by incorporating a utility function that has a trade-off between the information from the SLAM backend and the reduction of the map entropy. Stachniss et al. \cite{raobackwellStachniss} employ a Rao-Blackwellized particle filter (RBPF) for representing both the robot poses and the map. Through this, they can determine the best possible action by estimating the entropy of the particle filter. The use of Kullback-Leibler divergence as a metric for the measure of information can be found in \cite{kldluca}. Different optimality criterion from Theory of Optimal Design (TOED) \cite{pukelsheim2006optimal} is often used to quantify the uncertainity. For example, Carillo et al. \cite{carrillodopt} show that the D-optimality can be used as an information metric and is comparable to the A-optimality criterion. The recent approaches, especially the ones that incorporate graph-based optimization in SLAM have proven to have better long-term prediction than the RBPF approaches, which ultimately suffer from particle depletion as the size of the environment grows. \cite{DBLP:journals/corr/MuPALH15}

In this context, the presented work proposes a novel exploration strategy for Active SLAM devised with the dual objectives of sustaining an efficient exploration rate while optimizing SLAM accuracy. Our approach - denoted as \textit{"FIT-SLAM"} - an acronym representing 'Fisher Information and Traversability-Based Active SLAM' aims to address the two objectives. In detail, the 3D space exploration task is conceptualized as a 2D traversability map, taking into account the constraints posed by the 3D terrain and obstacles. This transformation results in a notable decrease in the time taken to find the frontier clusters as well as path generation to the candidate goals. Transforming the 3D exploration space to a 2D traversability map also eliminates the computational burden posed by maintaining a 3D Voxel map during exploration. Then, a goal selection approach taking into account the information gained upon reaching a given goal position, as well the information provided by the landmarks generated from SLAM during possible path execution towards this goal, is used to select the next destination point to be visited. The Fig. \ref{fig:enter-label} shows a simplified model of our approach. To the best of our knowledge, this is the first work for UGVs exploring a 3D environment where the strategy not only accounts for the safety of the robot during exploration but also provides promising results for the reduction of the uncertainty related to both the pose of the robot as well as the map generated by a SLAM backend. 

The main contributions of this work encompass two key aspects: (i) the proposition of frontier-based exploration and path planning strategies based on the use of a global traversability map and (ii) the proposition of a candidate goal selection approach which takes into account the information gained upon reaching the goal as well as the information provided by the landmarks generated from SLAM during the path execution to this goal.

\begin{figure}[t]
\input{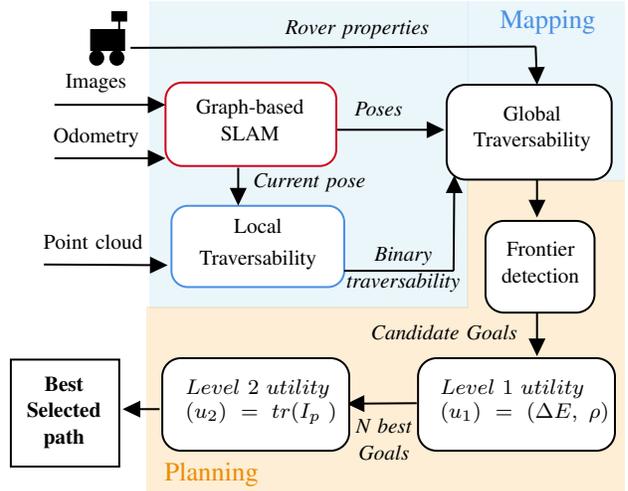}
\caption{Our ASLAM solution framework: First a global traversability map is built based on graph-based SLAM and 3D perception. Then, based on the rover capabilities, traversability scores are thresholded and the frontiers are detected. Goals are defined for each frontier and ranked on the basis of information gain. Finally, path safety for each goal are evaluated using predicted perception entropies. Depending on the other constraints of the mission, the final path is selected and executed.}
\label{fig:framework}
\end{figure}

The paper is organized as follows: sections \ref{sec:gbslam} and \ref{sec:trav} present the methodology used for the SLAM backend and the traversability estimation, respectively. The parts \ref{sec:frontier} and \ref{sec:utility} deal with the proposed frontier-based exploration method. The results are showcased in section \ref{sec:results}. Finally, conclusions and future work are discussed in section \ref{sec:conclu}.

\section{Methodology} 
\label{sec:methodology}

We propose a complete ASLAM framework to deal with the active exploration problem of an unknown and unstructured environment. Our solution is based on a graph-based SLAM approach for localization and a traversability estimator for path planning risk assessment. Finally, an analysis of the Fisher information calculated for each trajectory is used to select the best path. The proposed solution is summarized in Fig. \ref{fig:framework}.


\subsection{Graph-based SLAM}
\label{sec:gbslam}

The proposed navigation solution uses a graph-based SLAM approach in which nodes represent robot 6D poses $\mathbf{x}$ or landmarks $m_i$ of the map $\mathbf{M}$ poses, and edges in the map represent constraints between these poses. Such an approach makes use of a sliding window optimization that consists of finding the optimal state $\mathbf{X}^{*} = \{\mathbf{x}, \mathbf{M}\}$ that minimizes the summation of the norm of the residuals, that are the errors of all factors $\mathbf{e}$ weighted by their respective covariances $\Sigma_k$, such as:

\begin{equation}
    \mathbf{X}^{*} = \argmin_{\mathbf{X}}
    \left( \sum_{k \in \mathcal{G}} \norm{\mathbf{e}_k(\mathbf{X})}^2_{\Sigma_k} \right)
    \label{eq:optim}
\end{equation}
In the case of loop-closure detection, a similar optimization is performed but over the entire problem.

Note that for this work, we use RTABMap \cite{labbe2019rtab} as our SLAM backend. However, the proposed approach is not limited to RTABMap and can work with any other graph-based SLAM providing both key-frame poses and a map composed of landmarks.

\subsection{Traversability estimation}
\label{sec:trav}

Based on the pose-graph obtained by the SLAM approach, the detections obtained from the 3D LiDAR sensor (or other depth sensors) are registered in the world frame to process the traversability estimation. Following \cite{770402}, traversability illustrates the difficulty of navigating through a specific region and encompasses the suitability of the terrain for traversing based on its physical properties, such as slope, roughness and surface condition as well as the mechanical characteristics and capabilities of the UGV.

\begin{figure}[b]
    \begin{subfigure}{0.3\linewidth}
         \centering
         \includegraphics[height=3cm]{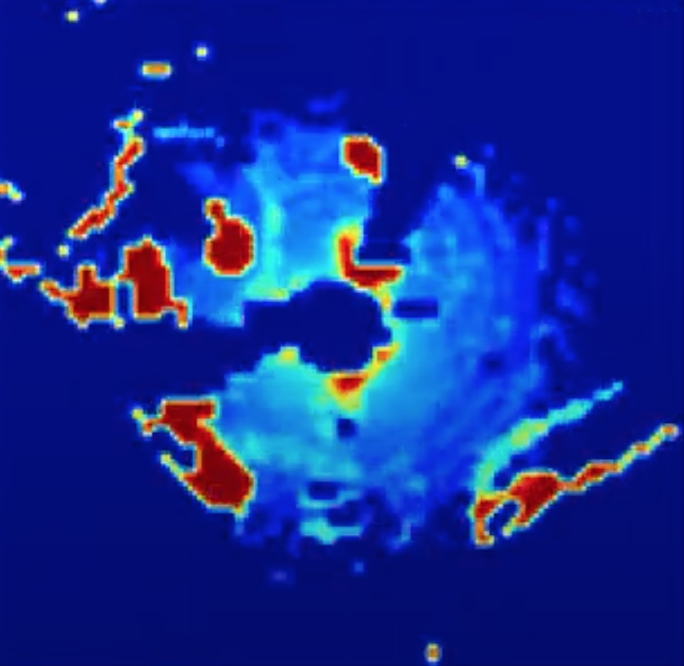}
         \caption{}
         \label{fig:traversability-map}
     \end{subfigure}
     \hfill
     \begin{subfigure}{0.64\linewidth}
         \centering
         \includegraphics[width=\textwidth]{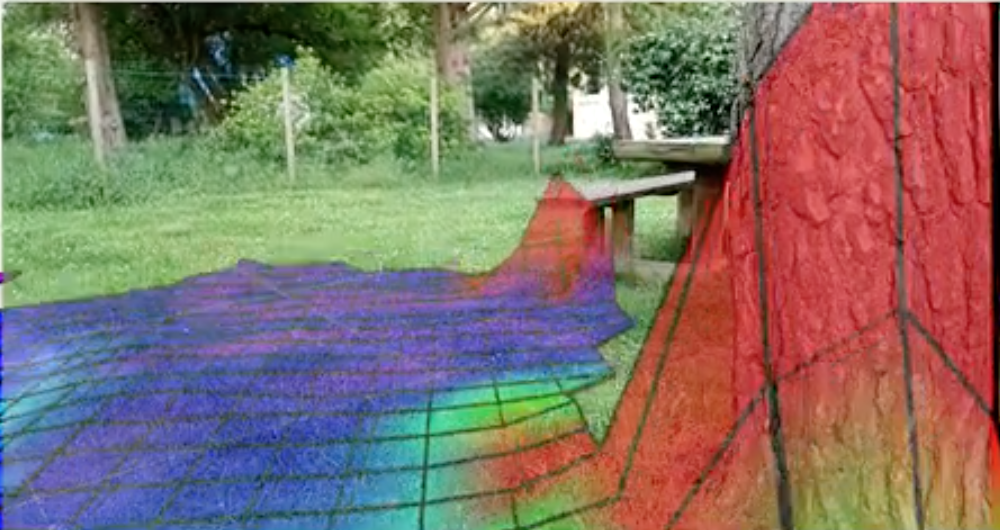}
         \caption{}
         \label{fig:traversability-img}
     \end{subfigure}    
    
    \caption{Estimated local traversability map. (a) The processed geometric traversability was obtained with a 3D LiDAR detection. (b) The reprojection of the traversability on the synchronized image showing (in red) the navigability risks.}
    \label{fig:traversability}
\end{figure}

In this work, we used a geometry-based traversability estimator similar to the works in \cite{annurevcontrol, cao2022autonomous, hudson2021heterogeneous}. The objective here is to model the environment as a grid of regularly spaced cells, 
where each cell typically represents the size of a UGV wheel. Each of these cells is populated with a traversability score or is tagged as unknown. This score is obtained by processing the statistics of the 3D points belonging to each cell, as in \cite{goldberg2002stereo}
An example of the local traversability obtained is presented in Fig. \ref{fig:traversability}. This transformation allows us to obtain a global georeferenced 2D map that represents the traversability of the 3D environment. Depending on the robot's risk acceptance criteria and capabilities, the traversability score can be \textit{thresholded} to generate a 2D binary traversability grid map. This grid map serves a function similar to that of an occupancy grid and is employed in path planning to maintain the robot's safety throughout its operations.

\subsection{Frontiers detection and clustering}
\label{sec:frontier}

\subsubsection{Frontiers detection}
Frontiers are regions in the map that act as a boundary between explored and unexplored regions. The proposed exploration algorithm begins with plotting an exploration boundary within which the frontiers are searched. A conventional frontier search algorithm similar to the one proposed in \cite{BYamuchi} is used.
    
\subsubsection{Frontiers clustering and candidate goal determination}
In our case, the conventional frontier search algorithm is modified to perform frontier clustering constrained to a maximum size. All linked frontier cells are treated together as one cluster, and the frontier point with the median index of the cluster is chosen as a candidate goal $C_i$. The path to reach each candidate goal is then generated. The 2D global traversability map coupled with the A* path planning algorithm from the ROS 2 navigation stack is used for this purpose.
In cases where it is impossible to calculate a path to the goal or when the goal has been previously identified as a frontier point in an earlier iteration, it is designated for exclusion in a blacklist, indicating that the goal is currently unreachable.
If there are no frontiers available in the map or in the area of interest, the exploration mission is considered to be a success and the mission ends.


\subsection{Utility computation and candidate goal selection}
\label{sec:utility}

The utility of any candidate goal $C_i$ is processed at two levels: The first-level $(u_1)$ is based on the computation of the total distance of the path from the current robot position to the candidate frontier and on the information gained upon reaching the said frontier. The second level $(u_2)$ is estimated by the information gained along the path.

\subsubsection{First-level utility computation $(u_1)$}
Let $C_x = \{C_0, C_1, \ldots, C_n\}$ be a set of candidate goals. For each candidate $C_i \in C_x$, we have an associated path $p_i$ with a length $\rho_i$ and a measure of information gained upon reaching the candidate goal $\Delta E_i$, which represents the change in map entropy upon observing the unknown cells after reaching the candidate goal.

More precisely, to compute the possible information gain upon reaching the candidate goal, we use a conventional ray tracing algorithm to get the set of potentially observable cells. Initially, we first determine the optimal arrival sensor orientation $\Theta^*_s$ that maximizes the information gain. In practice, to control the spatial density of the ray-tracing and expedite the process, we approximate $\Delta E_i$ using a finite number of rays separated by $\Delta\theta$. 

Let $\Theta = \{0, \Delta\theta, 2\Delta\theta, \cdots, 2\pi\}$ be the set of discretized ray directions and $\Phi$ is the camera field-of-view, we have the optimal arrival sensor orientation $\Theta_s^*$ given by:
\begin{equation}
     \Theta_s^* = \argmax_{\Theta_s \in \Theta} \left( \sum_{\Theta_s-\frac{\Phi}{2}}^{\Theta_s+\frac{\Phi}{2}} \Delta E^{\Theta_s}_G\right)
\end{equation}
where, $G = \{c_{1}, c_{2}, \cdots, c_{n}\}$ represents a 2D occupancy grid composed of $n$ cells.
$\Delta E^{\Theta_s}_G$ is the information gained along the ray in the direction $\Theta_s$. The information gain can be calculated as the change in the occupancy grid entropy before and after observing the cells along the ray in the direction $\Theta_s$.

Following \cite{bougsame}, the entropy $E^{\Theta_s}_G$ of a given occupancy grid-map $G$ is computed based on the \textit{Shannon entropy} as a measure of map uncertainty. It is given by:
\vspace{-2mm}
\begin{equation}
\begin{array}{r@{}l}
    E_{G}^{\Theta_s} = \displaystyle\sum_{i=0}^{n} &{} E^{\Theta_s}[c_i] 
    = -\displaystyle\sum_{i=0}^{n} \left(p(c_i) \cdot \log_2(p(c_i)) \right. \\
    &\quad\quad\quad\left. + (1 - p(c_i)) \cdot \log_2(1 - p(c_i)) \right)
\end{array}
\label{eq:entropy}
\end{equation}
where, $c_i$ represents a cell in $G$, and $p(c_i)$ represents the occupancy probability of the cell $c_i$. If the cell $c_i$ is unknown then $p(c_i) = 0.5$. It is manifest that such a computation requires the estimation of probabilities of all cells that can be observed along the ray. The farther a ray travels into unknown space, the more likely it is to be obstructed by an obstacle. Thus, as proposed in \cite{tNPotthast}, the observability of the cell is dependent on the previous cells traversed along the ray, such as:
\vspace{-2mm}
\begin{equation}
    p(x_r|x_{r-1}) =
    \begin{cases}
     1 & \text{if ray intersects an occupied cell} \\
     \gamma^N & \text{otherwise}
    \end{cases}
\end{equation}
where $p(x_r|x_{r-1})$ is the observability of a cell lying along the ray composed of $r$ cells, $\gamma$ is the degradation parameter which controls how fast the probability degrades along the ray, $N$ is the previous number of cells traversed by the ray. Given the observability of a cell, we can estimate the posterior occupancy probability of the cell $c_r$ as:
\begin{equation}
    p(c_r) = \frac{1 + p(x_r|x_{r-1})}{2}
    \label{eq:occprob}
\end{equation}

Finally, $\Delta E_i$, which is the information gained upon reaching the candidate goal $C_i$ can be processed with the optimal arrival sensor orientation $\Theta_s^*$ by:
\begin{equation}
        \label{eq:entropyfrontier}
        \Delta E_i = \sum_{\Theta_s^*-\frac{\Phi}{2}}^{\Theta^*_s+\frac{\Phi}{2}} \Delta E^{\Theta_s^*}_G
\end{equation}

At this step, the first-level utility $(u_1)$ can be computed for each candidate goal $C_i$ as a trade-off between the path length to be traversed by the robot $\rho_i$ and the information gained $\Delta E_i$ upon reaching the goal:
\begin{equation}
    u_1(C_i) =  \alpha ~N_{\rho^{-1}} ~{\rho_i^{-1}} + (1-~\alpha)~N_{\Delta E}~\Delta E_i
\end{equation}
where $\alpha \in [0,1]$ (resp. ($1-\alpha$)) represents the weight assigned to $\rho_i$ (resp. $\Delta E_i$) and $N_{\rho^{-1}}$, $N_{\Delta E}$ are normalization factors.
    
By ranking the set of candidate goals $C_x$ based on their corresponding first-level utility ($u_1$) values, we can determine the $N$ most promising candidate goals in terms of distance to the goal and overall map entropy reduction. 

However, this selection criterion does not take into account the information acquired during the travel to the intended goal. In instances where the information gathered during the travel phase is insufficient, the robot runs the risk of getting lost in the map and becoming disoriented. This could lead to a high localization uncertainty, which inherently could lead to a poor map accuracy. The inclusion of the second utility level mitigates this limitation.

\begin{figure}[b]
     \centering
     \includegraphics[height=4.5cm]{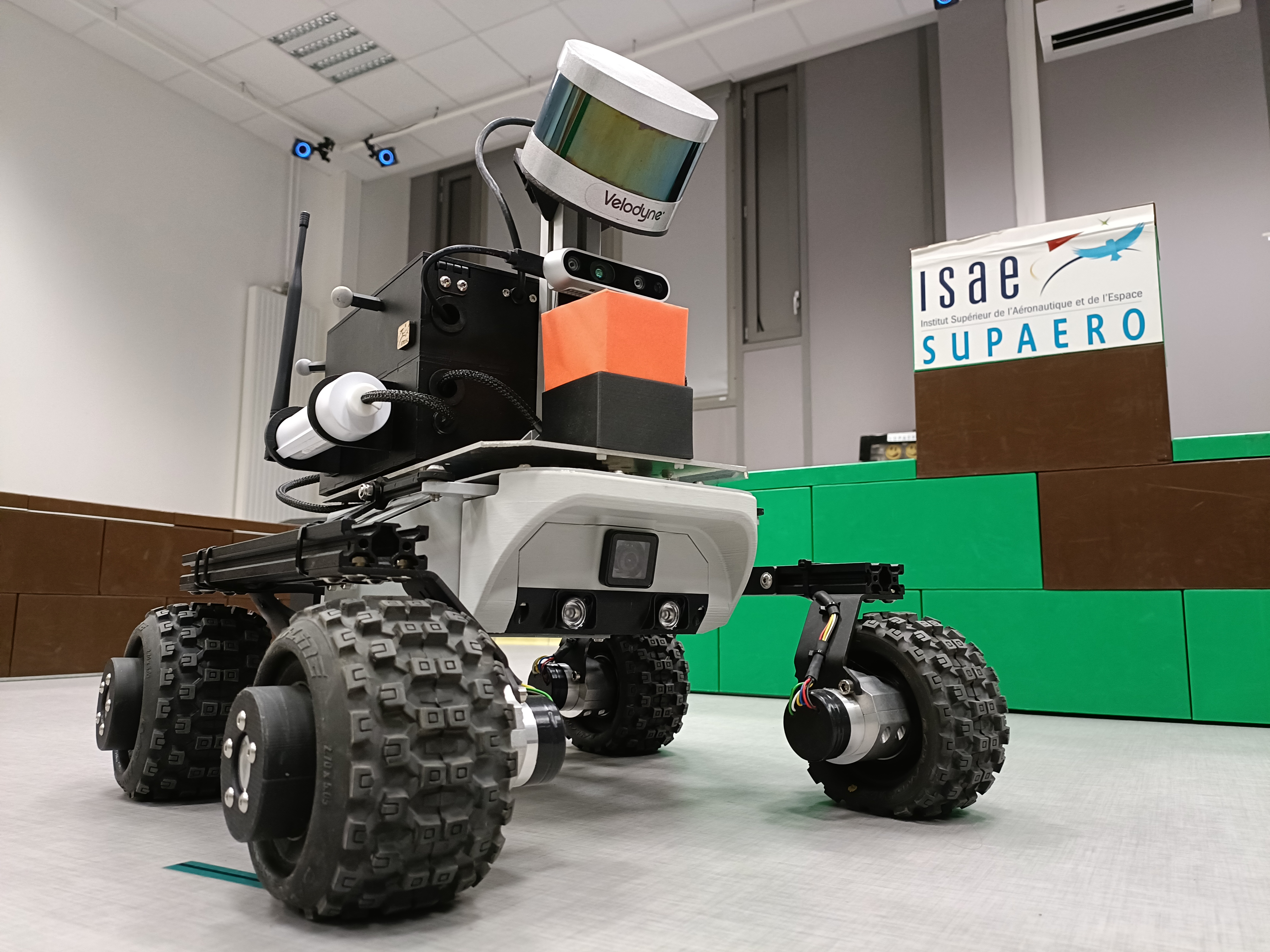}
     \caption{Our robotic platform equipped with the sensors (Depth Camera, 3D LiDAR, IMU and wheel odometry) required for the algorithm.}
    \label{fig:LeoRover}
\end{figure}

\subsubsection{Second level utility computation $(u_2)$}

We propose to add to the previous utility function the information gathered during path traversal. The objective is to maximize the robot's overall pose accuracy to the utmost degree and as a consequence, improve the map quality.

This level of utility computation will be used to determine the most optimal candidate goal by estimating the information gathered during the path execution. Given a path $p_i$ to a candidate goal $C_i$, we first sample the trajectory with a sampling distance equal to the max-depth of the camera FOV to obtain a set of waypoints $W_i = \{w_1, w_2, \ldots, w_n\}$. For each $w_k \in W_i$, we compute the information of all the landmarks from the map that lie within the FOV.
    
To do so, the Fisher Information Matrix (FIM) is used. FIM represents the minimum reachable covariance of an unbiased estimator \cite{fim_chen} and allows to quantify the estimation uncertainty. In our approach, we use the bearing vector representation of the 3D landmark. Given a 3D landmark that lies in a voxel $v_{w_i}$ in the world frame with a covariance $Q_i$, the observation function can be modeled using the bearing vector $b_i$ of the detection $v_{w_i}$ such as: 
    \begin{equation}
    b_i = \frac{v_{c_i}}{\|v_{c_i}\|_2} ~~\text{with}~~ v_{c_i} = T_{cw} v_{w_i}
    \label{eq:observation_model}
    \end{equation}
    
where $v_{c_i}$ is the $i^{th}$ voxel in the camera frame, $T_{cw}$ the affine transform matrix from world to camera frame.

The Fisher information matrix $I_i$ corresponding to the considered landmark lying in the voxel $v_{w_i}$ can be derived as:
    \begin{equation}
        I_i = J_i ~ Q_i ~ J_i^{T}
        \label{eq:FIM}
    \end{equation}

With $J_i$ the jacobian of the observation model (\ref{eq:observation_model}) given by:
    \begin{equation}
        J_i = \frac{\partial b_i}{\partial T_{wc}} = \frac{\partial b_i}{\partial v_{c_i}} \frac{\partial v_{c_i}}{\partial T_{wc}}
        \label{eq:firstjacob}
    \end{equation}
where the elements in Eq. (\ref{eq:firstjacob}) are given using the Special Euclidean group SE(3) \cite{beyondpcl} by:
\begin{equation}
\frac{\partial b_i}{\partial v_{c_i}} = \frac{1}{\|v_{c_i}\|_2} I_3 - \frac{v_{c_i} (v_{c_i})^T}{(\|v_{c_i}\|_2)^3}
\end{equation}
\begin{equation}
        \frac{\partial v_{c_i}}{\partial T_{wc}} = R_{cw} \left[ -I_3, \left[v_{w_i} \right]_\times \right]
\end{equation}
    
where $R_{cw}$ is the rotation from world to camera frame and $[v]_\times$ is the cross-matrix of vector $v$.
    
Storage of the FIMs for all the voxels arguably consumes a lot of memory. A common way to bypass this memory usage is to use the theory of optimal experimental design (TOED) \cite{pukelsheim2006optimal}, which utilizes the T-opt optimality criterion, i.e., the trace of the FIM can be used to convert FIM to a scalar value significantly reducing the memory usage.

\begin{figure}[t]
\centering
\begin{minipage}{.48\columnwidth}
\hspace*{-2ex}
  \centering
    \begingroup
      \tikzset{every picture/.style={scale=0.84}}%
      \begin{tikzpicture}[yscale=0.75,xscale=0.75]
        \begin{axis}[
            xlabel={Time (s)},
            ylabel={tr(Covariance)},
            grid=major,
            xticklabel style={/pgf/number format/fixed,
            /pgf/number format/1000 sep={}},
            yticklabel style={/pgf/number format/fixed,
            /pgf/number format/1000 sep={}},
            legend entries={FIT-SLAM, Greedy frontier selection, Random frontier selection},
            legend style={at={(1.17,0.39)},anchor=north east},
            xmin=0,  
            xmax=4700,  
            ymin=0,  
            ymax=25,  
        ]
        \addplot+ [mark = none, line width=1.5pt] table [x=Time, y=Covariance, col sep=comma] {Experiment-Data/ours_loc_cov.csv};
        \addlegendentry{FIT-SLAM}
        
        \addplot+ [mark = none, line width=1.5pt] table [x=Time, y=Covariance, col sep=comma] {Experiment-Data/greedy_loc_cov.csv};
        \addlegendentry{Greedy}
        
        \addplot+ [mark=none, line width=1.5pt] table [x =Time, y=Covariance, col sep=comma] {Experiment-Data/random_loc_cov.csv};
        \addlegendentry{Random}

        \end{axis}
    \end{tikzpicture}
    \endgroup
    \vspace{-3ex}
  \caption*{(a)}
\end{minipage}%
\begin{minipage}{.48\columnwidth}
  \centering
    \begingroup
      \tikzset{every picture/.style={scale=0.84}}%
          \begin{tikzpicture}[yscale=0.75,xscale=0.75]
        \begin{axis}[
            xlabel={Time (s)},
            ylabel={\% unexplored Map},
            grid=major,
            xticklabel style={/pgf/number format/fixed,
            /pgf/number format/1000 sep={}},
            yticklabel style={/pgf/number format/fixed,
            /pgf/number format/1000 sep={}},
            legend entries={FIT-SLAM, Greedy frontier selection, Random frontier selection},
            legend style={at={(0.56,0.39)},anchor=north east},
            xmin=0,  
            xmax=4900,  
            ymin=0,  
            ymax=100,  
        ]
        
        \addplot+ [mark=none, line width=1.5pt] table [x=Time, y=MapData, col sep=comma] {Experiment-Data/ours_map_data_coverage.csv};
        \addlegendentry{FIT-SLAM}
        \addplot+ [mark=none, line width=1.5pt] table [x=Time, y=MapData, col sep=comma] {Experiment-Data/greedy_map_data_coverage.csv};
        \addlegendentry{Greedy}
        \addplot+ [mark=none, line width=1.5pt] table [x=Time, y=MapData, col sep=comma] {Experiment-Data/random_map_data_coverage.csv};
        \addlegendentry{Random}
        \end{axis}  
        \end{tikzpicture}
    \endgroup  
    \vspace{-3ex}
  \caption*{(b)}
\end{minipage}
\caption{Evaluation of the proposed approach for the experiment conducted in simulation (a) Evolution of the trace of the robot state covariance over time. (b) Evolution of the exploration rate}
\label{fig:explocov-vs-time}
\end{figure}

Therefore, we can estimate the information of the path $p_i$ by summing over all waypoints $w_k \in W_i$.
    \begin{equation}
        I_{p_i} = N_I \sum_{w_k \in w_x} I_{w_k} = N_I \sum_{w_k \in w_x} \sum_{~v_{w_i} \in v_{w_k}} I_i
    \end{equation}
where $I_{p_i}$ is the information of the path, $I_i$ is the information of the voxel $v_{w_i}$ and $N_I$ is the normalization factor. However, it is important to note that even though the memory usage is significantly reduced due to the inclusion of T-opt criteria, the computation of the FIM, even with the incorporation of voxelization is a computationally intensive task. 


Following the computation of information along the path, we are able to compute the second level utility ($u_2$) for a candidate goal $C_i$, such as:
    \begin{equation}
        u_2(C_i)= \beta \cdot u_1 + ~(1-\beta) I_{p_i}
    \end{equation}
where $\beta \in [0,1]$ is a weighting parameter between $u_1$ and $I_{p_i}$. We compute this information solely for the $N$ best candidate goals ranked after the computation of $u_1$. Let $C_x^*$ represents the $N$ best candidate goals in $C_x$, the best candidate goal $C_{best}$ is simply the one with the biggest utility value ($u_2$), such as:
\begin{equation}
        C_{best}= \argmax_{C_i \in C_x^*} \left (\beta \cdot u_1 + ~(1-\beta) I_{p_i} \right)
    \end{equation}
Finally, the path selected based on $u_{2}$ is the one that minimizes the localization and map uncertainty as it has the most informative landmarks observed during traversal while also taking into account the distance of the goal and information gained upon reaching the said goal.

\section{Results}
\label{sec:results}
We validated our approach by comparing two metrics: (i) the percentage of unexplored map with respect to time and (ii) the evolution of the localization covariance (the marginal error obtained after graph optimization from the SLAM) with respect to time during exploration.
The entirety of our system is programmed using the ROS 2 framework and tested in a 3D unstructured simulated environment on Gazebo running on a standard computer. Our real-world experiment was conducted in a planar environment with the Nvidia Jetson AGX Xavier (CPU: 8 Core @ 2.26 GHz, RAM: 32 GB, GPU: unused) as our on-board computer and the LeoRover as our robotic platform. The Intel Realsense D435 and the Velodyne VLP-16 LiDAR was used as our primary sensors. The parameters used for our experiments are, $\alpha =0.35, ~\beta=0.4, ~\Delta\theta=8.5^\circ$. $N=7$, traversability map resolution = $0.05m$ and the voxel size is $0.25m$. 






\begin{figure}[t]
\centering
\begin{minipage}{.48\columnwidth}
\hspace*{-2ex}
  \centering
    \begingroup
      \tikzset{every picture/.style={scale=0.84}}%
      \begin{tikzpicture}[yscale=0.75,xscale=0.75]
        \begin{axis}[
            xlabel={Time (s)},
            ylabel={tr(Covariance)},
            grid=major,
            xticklabel style={/pgf/number format/fixed,
            /pgf/number format/1000 sep={}},
            yticklabel style={/pgf/number format/fixed,
            /pgf/number format/1000 sep={}},
            legend entries={FIT-SLAM, Greedy frontier selection, Random frontier selection},
            legend style={at={(0.57,1.15)},anchor=north east},
            xmin=0,  
            xmax=3200,  
            ymin=0,  
            ymax=2.5,  
            xtick={0,1000,2000,3000},
        ]
        \addplot+ [mark = none, line width=1.5pt] table [x=Time, y=Covariance, col sep=comma] {Experiment-Data/LeoTest_ours_loc_cov.csv};
        \addlegendentry{FIT-SLAM}
        
        \addplot+ [mark = none, line width=1.5pt] table [x=Time, y=Covariance, col sep=comma] {Experiment-Data/LeoTest_greedy_loc_cov.csv};
        \addlegendentry{Greedy}
        
        \addplot+ [mark=none, line width=1.5pt] table [x =Time, y=Covariance, col sep=comma] {Experiment-Data/LeoTest_random_loc_cov.csv};
        \addlegendentry{Random}

        \end{axis}
    \end{tikzpicture}
    \endgroup
    \vspace{-3ex}
  \caption*{(a)}
\end{minipage}%
\begin{minipage}{.48\columnwidth}
  \centering
    \begingroup
      \tikzset{every picture/.style={scale=0.84}}%
          \begin{tikzpicture}[yscale=0.75,xscale=0.75]
        \begin{axis}[
            xlabel={Time (s)},
            ylabel={\% unexplored Map},
            grid=major,
            xticklabel style={/pgf/number format/fixed,
            /pgf/number format/1000 sep={}},
            yticklabel style={/pgf/number format/fixed,
            /pgf/number format/1000 sep={}},
            legend entries={FIT-SLAM, Greedy frontier selection, Random frontier selection},
            legend style={at={(0.56,0.39)},anchor=north east},
            xmin=0,  
            xmax=3200,  
            ymin=0,  
            ymax=100,  
            xtick={0,1000,2000,3000},
        ]
        
        \addplot+ [mark=none, line width=1.5pt] table [x=Time, y=MapData, col sep=comma] {Experiment-Data/LeoTest_ours_map_data_coverage.csv};
        \addlegendentry{FIT-SLAM}
        \addplot+ [mark=none, line width=1.5pt] table [x=Time, y=MapData, col sep=comma] {Experiment-Data/LeoTest_greedy_map_data_coverage.csv};
        \addlegendentry{Greedy}
        \addplot+ [mark=none, line width=1.5pt] table [x=Time, y=MapData, col sep=comma] {Experiment-Data/LeoTest_random_map_data_coverage.csv};
        \addlegendentry{Random}
        \end{axis}  
        \end{tikzpicture}
    \endgroup  
    \vspace{-3ex}
  \caption*{(b)}
\end{minipage}
\caption{Evaluation of the proposed approach for the real-world experiment. (a) Evolution of the trace of the robot state covariance over time. (b) Evolution of the exploration rate. The jumps in the covariance trace correspond to the loop closure detections.}
\label{fig:explocov-vs-time-leo}
\vspace{-6mm}
\end{figure}
Results for our experiments in simulation and real-world are shown in Fig. \ref{fig:explocov-vs-time} and Fig. \ref{fig:explocov-vs-time-leo}. Three frontier-based exploration methods have been tested. (i) Our approach (FIT-SLAM) which uses the traversability estimation coupled with the most informative path choice. (ii) A random selection of the frontier. (iii) A greedy frontier selection approach which selects the closest frontier to the robot. It is important to note that the very same 2D traversability map is used to generate paths and detect frontiers in all the three approaches. 

The plot in
Fig. \ref{fig:explocov-vs-time}~(a) shows that our approach has the best accuracy. In Fig. \ref{fig:explocov-vs-time-leo}~(a), several drops in covariance can be observed. These indicate the points of loop-closure. The high number of loop closures detected in our method is a direct consequence of using the information provided by the landmarks of the map in the goal selection. Regarding exploration speed results, illustrated in Fig. \ref{fig:explocov-vs-time}~(b) and \ref{fig:explocov-vs-time-leo}~(b), it is interesting to observe that the greedy frontier approach maps the environment very quickly in the initial phase. However, during long-term planning, the greedy frontier exploration is stuck exploring frontiers that yield very low information gain. This result suggests that our first-level utility would play a key role in exploration speed improvement. The results for the real-world experiment are consistent with the results of the simulation.

\section{Conclusion and future work}
\label{sec:conclu}
We proposed a novel Active-SLAM approach to explore a 3D unstructured environment based on a traversability map. Our solution uses both the Shannon entropy to measure the amount of information that could be gained by mapping new areas and the Fisher information matrix as an information metric to estimate the information gained during path execution by observing the known and mapped landmarks. Our entire approach has been tested in a 3D setting on simulation and we showed substantial improvements in the exploration rate and accuracy of the SLAM. We also validated the solution with a real-time experiment with a real robot in a controlled 2D environment. In the near future, we plan to extend our real-world experiment in a 3D environment and also compare our algorithm against other existing methodologies. We also aim to extend our approach to multi-robot exploration missions.

\bibliographystyle{unsrt}
\bibliography{bib}

\vspace{12pt}

\end{document}